\documentclass[final]{cvpr}

\usepackage{times}
\usepackage{epsfig}
\usepackage{graphicx}
\usepackage{amsmath}
\usepackage{amssymb}

\usepackage{caption}
\usepackage{subcaption}
\usepackage{multirow}
\usepackage{algorithm}
\usepackage[noend]{algpseudocode}

\DeclareMathOperator*{\argmin}{arg\,min}

\usepackage[pagebackref=true,breaklinks=true,colorlinks,bookmarks=false]{hyperref}

\def\cvprPaperID{47} 
\def\confYear{CVPR 2021}

\begin{document}

\title{A Bop and Beyond: A Second Order Optimizer for Binarized Neural Networks\thanks{Preprint accepted to the LatinX in CV Research Workshop at CVPR'21}}

\author{Cuauhtemoc Daniel Suarez-Ramirez, Miguel Gonzalez-Mendoza, Leonardo Chang-Fernandez, \\
Gilberto Ochoa-Ruiz, Mario Alberto Duran-Vega\\
Department of Computer Science\\
Tecnologico de Monterrey, School of Engineering and Sciences\\
Monterrey, NL  \\
{\tt\small a01206503@exatec.tec.mx, mgonza@tec.mx, lchang@tec.mx,}\\
{\tt\small  gilberto.ochoa@tec.mx, a00755076@exatec.tec.mx}
}

\maketitle

\begin{abstract}
   The optimization of Binary Neural Networks (BNNs) relies on approximating the real-valued weights with their binarized representations. Current techniques for weight-updating use the same approaches as traditional Neural Networks (NNs) with the extra requirement of using an approximation to the derivative of the sign function - as it is the Dirac-Delta function - for back-propagation; thus, efforts are focused adapting full-precision techniques to work on BNNs. In the literature, only one previous effort has tackled the problem of directly training the BNNs with bit-flips by using the first raw moment estimate of the gradients and comparing it against a threshold for deciding when to flip a weight (Bop). In this paper, we take an approach parallel to Adam which also uses the second raw moment estimate to normalize the first raw moment before doing the comparison with the threshold, we call this method Bop2ndOrder. We present two versions of the proposed optimizer: a biased one and a bias-corrected one, each with its own applications. Also, we present a complete ablation study of the hyperparameters space, as well as the effect of using schedulers on each of them. For these studies, we tested the optimizer in CIFAR10 using the BinaryNet architecture. Also, we tested it in ImageNet 2012 with the XnorNet and BiRealNet architectures for accuracy. In both datasets our approach proved to converge faster, was robust to changes of the hyperparameters, and achieved better accuracy values.
\end{abstract}

\section{Introduction}

Artificial Intelligence (AI) is having a great momentum in terms of new applications and positive impacts on society. Deep Learning (DL) is a sub-area of AI that has demonstrated outstanding capabilities for solving complex tasks in many areas. In computer vision in particular, it has out-performed previous approaches in tasks such as Image Classification, Image Recognition, and Image Segmentation\cite{Lecun2015deep, Goodfellow2016deep, Schmidhuber2015deep, Redmon2018yolov3, Loyola2020review}. 


Current approaches for optimizing DL methods (i.e. for edge computing applications) are either based on constructing and training lighter neural networks or pruning larger ones. In particular, in regard to lighter neural networks, approaches based on Binarized Neural Networks (BNNs), which uses weights constrained to $\{-1, +1\}$, result in models which are much less computationally expensive, and lead to noticeable reductions in energy consumption when implemented on specialized hardware, and far less memory usage. Moreover, the nature of the BNNs impacts other types of applications such as for Neuromorphic Computing \cite{Krestinskaya2018binary, Lammie2019variationaware, Srinivasan2019restocnet, Lu2020spiking} and Quantum Computing \cite{Fawaz2019quantum}, thus, highlighting the importance of these type of networks.

Anderson and Berg \cite{Alex2017highdimensional} proved theoretically and experimentally that BNNs maintain the geometrical properties of the Convolutional Neural Networks (CNNs), specifically the properties of the convolution operation. This means that the angle between a stochastic vector and its binarized counter-part converges to a small value with an increasing number of dimensions. Also, the matrix-product is preserved.

BinaryNet \cite{Courbariaux2016binarized} was the pioneering work proving that BNNs are viable for complex tasks such as image classification on ImageNet \cite{ILSVRC15, Deng2009imagenet}. XnorNet \cite{Rastegari2016xnornet} proved that the binarized convolution operation could be done by just using \texttt{xnor} and \texttt{pop-count} operations which dramatically reduces the convolution's complexity. Since then, the subject has attracted attention and various papers have been published for reducing the loss function and training/validation errors, training deeper BNNs, and using multiple binary bases for the matter \cite{Liu2018bireal, Darabi2018regularized, Darabi2019bnn, Deveci2018energy, Lahoud2019selfbinarizing, Peters2018probabilistic, Bethge2019simplicity}.

Even though BNNs have been greatly improved, the methods for training them have remained mostly unchanged; they use Stochastic Gradient Descent or an equivalent method. Helwegen, \etal \cite{Helwegen2019latent} proposed a Binary Optimizer (Bop) which instead of using the gradients to update a ``latent weight", it uses this information to determine when to ``flip" the weights, directly training the net with 0s and 1s. This method calculates a raw average of the gradients (first raw moment), and compares it to a threshold to assess when to modify the weight.

The main focus of this paper is to implement an optimizer which only takes into account when to flip the weights sign/values inspired by Adam \cite{kingma2014adam} instead of using adapted full-precision methods. Our contributions are:

\begin{enumerate}
    \item Introduce a second order optimizer for BNNs which uses the first and second momentum of the gradients. This optimizer yields better results in terms of accuracy (tested on CIFAR-10 and ImageNet 2012) than the start-of-the-art methods including the first order optimizer Bop.
    \item Explore the effects of each hyperparameter, and study the effects of using schedulers on them.
\end{enumerate}

With this paper we introduce a specialized optimizer for BNNs which uses the first and second raw moment estimates of gradients to assess when to modify the sign of the binarized weight. This method is based on Bop while introducing the advantages of Adam. The obtained results outperform the other learning methods for BNNs.

\section{Background}
Consider a neural network, $y = f(x, w)$, with $w \in \mathbb{R}^n$, and a defined loss function $L(y, y_{label})$, where $y_{label}$ is the ground truth (real label). Then, the binarization problem is defined as:

\begin{equation} \label{eq:Binary_problem}
    w_{bin}^* = \argmin_{w_{bin} \in \{ -1, +1 \}^n} {\mathbb{E}_{x,y}{[L(f(x, w_{bin}), y_{label})]} }
\end{equation}

As global optimums usually cannot be found, approximate solutions via Stochastic Gradient-Descent (SGD) are used instead. The problem arises when evaluating the gradient $\frac{\partial L}{\partial w}$ depends on $\frac{\partial w_{bin}}{\partial w}$. During the forward pass, the binarization of the inputs and weights is achieved by using the $sign$ function:

\begin{equation} \label{eq:Binarization}
    w_{bin} = sign(w)
\end{equation}

\begin{figure}
  \centering
  \includegraphics[width=0.8\linewidth]{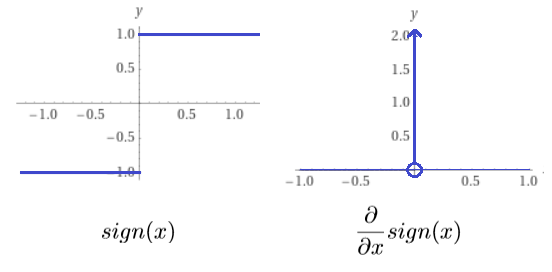}
  \caption{Sign function and its derivative, the Dirac delta function. The derivative banishes everywhere but in 0.}
  \label{fig:sign_derivative}
\end{figure}

As the gradient $sign$ function is the Dirac Delta function, it vanishes on every point except on $0$ (as seen in Figure \ref{fig:sign_derivative}). Thus, approximations must be used for calculating the derivative and use it for the gradients of the weights.

One of the most common approximations is the Straight-Through Estimator (STE) \cite{Yin2019understanding} which is defined as:

\begin{equation} \label{eq:STE}
    \frac{\partial L}{\partial w} = \frac{\partial L}{\partial w_{bin}}1_{\| w \| \leq t_{clip}}
\end{equation}

In other words, it permits the gradient to go through, except for those values where the weights have a large magnitude. The most common case is to use $t_{clip} = 1$, but Bethge, \etal \cite{Bethge2019simplicity} tested various values and found that values between $1.25$ and $1.5$ work better.

The first approaches towards improving the optimization methods for BNNs used different approximations to the sign function so it could be used for back-propagation. Some  of these approaches use a first or second order polynomial approximation \cite{Liu2018bireal} to the sign function or extra mathematical operations to consider the magnitude of the weight or the gradient. Still, the most common approximation is the STE or Clip \cite{Yin2019understanding, Rastegari2016xnornet, Courbariaux2015binaryconnect, Courbariaux2016binarized, Qin2020survey, Simons2019review} and the ApproxSign \cite{Liu2018bireal}. Recently, Bethge \etal \cite{Bethge2019simplicity} proved that there is no need for using a different approximation to STE, but the threshold must be tuned with values different from 1. 

\subsection{Real-valued approximation with binarized weights}


The most common algorithm for training BNNs is described in \ref{alg:TradBNNs}.

\begin{algorithm}
    \caption{Traditional training algorithm for BNNs using Stochastic-Gradient Descent on latent weights as presented by Helwegen \etal \cite{Helwegen2019latent}}\label{alg:TradBNNs}
    \begin{algorithmic}[1]
        \State \textbf{input:} Loss function $L(f(x, w), y)$, Batch size $K$
        \State \textbf{input:} Optimizer $\mathcal{P}: g \mapsto \delta_w$, learning rate $\alpha$
        \State \textbf{input:} Pseudo-Gradient $\Phi : L(f(x, w), y) \mapsto g \in \mathbb{R}^n$
        \State initialize $w \leftarrow w_0 \in \mathbb{R}^n$
        \While{stopping criterion not met}
            \State Sample mini-batch $\{x^{(1)}, \dots, x^{(K)}\}$;
            \State Gradient: $g \leftarrow \frac{1}{K} \Phi \sum_k{L\left( f(x^{(k)}; w), y^{(k)} \right)}$;
            \State Update latent weights: $w \leftarrow w + \alpha \mathcal{P}(g)$;
        \EndWhile
        
    \end{algorithmic}
    \label{alg:optRealWeights}
\end{algorithm}

Conceptually, this seems as a flawed approach as the only aspect that is enforced is to find when to change the magnitude of the weight, requiring the use of some pseudo-weight to be optimized and then binarized. One of the problems of using this approach is that the gradient with respect to the binary weight is not enough to trigger the change of its sign \cite{Liu2018bireal}. Therefore, a magnitude-aware gradient was proposed by Liu \etal which takes into account the magnitude of the gradient for changing the weights.

The problem of using ``latent weights" as a way of storing a non-real value by proxy that will be optimized upon for obtaining the real binary weight, was tackled by Helwegen \etal . They demonstrated in their paper that the binarization problem could actually be seen as the sign of the weight multiplied by its magnitude (momentum). Thus, latent weights were not needed as the problem was almost identical to using momentum for training full-precision weights.

\section{Related Work}
There has been various attempts to adapt full-precision optimizers to BNNs, but only one that directly tackles the problem of building a specialized one, Binary Optimizer (Bop), proposed by Helwegen \etal \cite{Helwegen2019latent}. This method is based on the concepts of momentum, and more notably, it showed that only flipping the sign of the weights is enough for correctly binarized the weights.


If the sign of the weights is all that matters, an optimization algorithm that only decides when to flip values makes a more natural manner to learn the weights. Helwegen \etal \cite{Helwegen2019latent} realized this, and proposed the idea to design a momentum-like algorithm that stores the gradients of each weight (with an exponential decay) and when the accumulated value surpasses a threshold, the weight flips. This function is shown in algorithm \ref{alg:Bop}.

\begin{algorithm}
    \caption{Bop, an optimizer for BNNs \cite{Helwegen2019latent}. It uses the first raw moment estimate to decide when to flip the weight value}\label{alg:Bop}
    \begin{algorithmic}[1]
        \State \textbf{input:} Loss function $L(f(x, w), y)$, Batch size $K$
        \State \textbf{input:} Threshold $\tau$, adaptivity rate $\gamma$;
        \State initialize $w \leftarrow w_0 \in \{ -1, 1 \}^n, m \leftarrow m_0 \in \mathbb{R}^n$
        \While{stopping criterion not met}
            \State Sample mini-batch $\{x^{(1)}, \dots, x^{(K)}\}$;
            \State Gradient: $g \leftarrow \frac{1}{K} \frac{\partial L}{\partial w} \sum_k{L\left( f(x^{(k)}; w), y^{(k)} \right)}$;
            \State Update momentum: $m \leftarrow (1 - \gamma)m + \gamma g$;
            \For{$i \leftarrow 1$ \textbf{to} n}
                \If{$\| m_i \| > \tau$ and $sign{(m_i)} = sign{(w_i)}$}
                    \State $w_i \leftarrow -w_i$;
                \EndIf
            \EndFor
        \EndWhile
        
    \end{algorithmic}
\end{algorithm}

The latent weights are better understood when thinking of the magnitude and the sign separately:

\begin{align}
\label{eq:latentW}
\begin{split}
 \tilde{w} = sign(w) \cdot \|w\| := w_{bin} \cdot m,
\\
 w_{bin} \in \{-1, +1\}, \quad m \in \left[ 0, \infty \right)
\end{split}
\end{align}

The main rationale for this approach is that the latent weights encode the inertia values $m$ of the binary weights $w_{bin}$. The bigger the magnitude is, the stronger the effect of this binarized weight. Thus, the gradient can be seen as an indicator of the ``necessity" of changing the sign of the weight.

\section{Second Order Binary Optimizer (Bop2ndOrder)}
As determining when to change the value of a given weight (when to flip it) is the main purpose of the optimization step, our optimizer uses the gradient for this purpose, as outlined above. For this, we analyzed some full-precision optimizers and we found a way to adapt them to obtain this information. The Adam optimizer \cite{kingma2014adam} seemed as the best choice to be adapted both conceptually and practically, due to two characteristics: 1) it resembles the work done in Bop \cite{Helwegen2019latent}, and 2) it further improves it by also using the second raw moment estimate.

As mentioned above, Bop is a first order binarized optimization method that calculates the first raw moment estimate of the gradients in the following way:

\begin{equation} \label{eq:Mt_Bop}
    m_t = (1 - \gamma)m_{t - 1} + \gamma g_t = \sigma \sum_{r=0}^{t}{(1 - \gamma)^{t - r}g}
\end{equation}

Then, this value is compared against a pre-defined threshold for deciding when to flip the weight:

\begin{equation} \label{eq:Bop_cases}
    w_t^i =
  \begin{cases}
    -w_{t-1}^i       & \quad \text{if } \| m_t^i \| \geq \tau \text{ and } \text{sign}(m_t^i) = \text{sign}(w_{t-1}^i)\\
    w_{t-1}^i        & \quad \text{otherwise.}
  \end{cases}
\end{equation}

One thing to note, is that not only the absolute value of $m_t$ should be higher than the threshold, but the sign of it should be the same as the previous weight. This is due to the gradient indicating the way of the maximum rate of change; if the weight points already in that direction, there is no reason to flip it again.

The natural iteration over this algorithm, is to also include a second raw moment estimate value for regularizing the gradients in the form of:

\begin{equation} \label{eq:Vt_Bop2ndOrder}
    v_t = (1 - \sigma)v_{t - 1} + \sigma g_t^2 = \sigma \sum_{r=0}^{t}{(1 - \sigma)^{t - r}g^2}
\end{equation}

As inspired by Adam, this would normalize the gradient making it invariant to re-scaling, and would make the training smoother and faster (in terms of the number of iterations). Thus, the previous quantity $m_t$ is converted into:

\begin{equation} \label{eq:St_Bop2ndOrder}
    s_t = \frac{m_t}{\sqrt{v_t} + \epsilon} 
\end{equation}

Transforming the comparison rule into:

\begin{equation} \label{eq:Bop2ndOrder_cases}
    w_t^i =
      \begin{cases}
        -w_{t-1}^i       & \quad \text{if } \| s_t^i \| \geq \tau \text{ and } \text{sign}(s_t^i) = \text{sign}(w_{t-1}^i)\\
        w_{t-1}^i        & \quad \text{otherwise.}
  \end{cases}
\end{equation}

This iteration, a novelty of our method is shown in algorithm \ref{alg:2ndBop}. If we compare this approach to Adam \cite{kingma2014adam}, we notice that no method for correcting bias is presented in the latter. Although, according to the results of Helwegen \etal \cite{Helwegen2019latent} this is not really necessary as the threshold value takes on this role through schedulers. However,  we decided to include an unbiased version of the algorithm as it would mitigate overfitting by reducing the bias of the mean and variance raw estimates. This is done by changing \eqref{eq:St_Bop2ndOrder} to:

\begin{equation} \label{Unbiased_St}
     s_t = \frac{m_t / \gamma}{\sqrt{v_t / \sigma} + \epsilon} 
\end{equation}

\begin{algorithm}
    \caption{Second Order Bop. This algorithm uses both the first and second raw moment estimates. Depending on if we choose the biased or unbiased algorithm, is the value of $s_t$ calculated.}\label{alg:2ndBop}
    \begin{algorithmic}[1]
        \State \textbf{input:} Loss function $L(f(x, w), y)$, Batch size $K$
        \State \textbf{input:} Threshold $\tau$, adaptivity rate $\gamma$, standard rate $\sigma$;
        \State initialize $w \leftarrow w_0 \in \{ -1, 1 \}^n, m \leftarrow m_0 \in \mathbb{R}^n, v \leftarrow v_0 \in \mathbb{R}^n$
        \While{stopping criterion not met}
            \State Sample mini-batch $\{x^{(1)}, \dots, x^{(K)}\}$;
            \State Gradient: $g \leftarrow \frac{1}{K} \frac{\partial L}{\partial w} \sum_k{L\left( f(x^{(k)}; w), y^{(k)} \right)}$;
            \State Update momentum: $m \leftarrow (1 - \gamma)m + \gamma g$;
            \State Update raw variance: $v \leftarrow (1 - \sigma)v + \sigma g^2$;
            \State Standardized momentum: $s \leftarrow s\_value(m, v)$;
            \For{$i \leftarrow 1$ \textbf{to} n}
                \If{$\| s_i \| > \tau$ and $sign{(s_i)} = sign{(w_i)}$}
                    \State $w_i \leftarrow -w_i$;
                \EndIf
            \EndFor
        \EndWhile
        
        \Function{$s\_value$}{$m, v$}
            \If{biased}
                \State \Return $\frac{m}{\sqrt{v} + \epsilon}$
            \Else
                \State \Return $\frac{m / \gamma}{\sqrt{v / \sigma} + \epsilon}$
            \EndIf
        \EndFunction
        
    \end{algorithmic}
\end{algorithm}

\section{Experimental Results}
In this section, we tested thoroughly the behavior of the Bop2ndOrder algorithm (both biased and unbiased) exploring the effects of its three hyperparameters ($\gamma, \sigma$ and $\tau$) and the effect  of scheduling policies (both increasing and decreasing the values) to those hyperparameters. Additionally, we tested whether the batch or layer normalization works best for our optimizer. Lastly, we compared our results against those obtained for Bop.

For the hyperparameters exploration, we used the BinaryNet \cite{Courbariaux2016binarized} architecture and tested the code in Google Colab using either Tesla P-100 or V-100 GPUs.

We included the metric of $\pi_t$ introduced by Helwegen \etal \cite{Helwegen2019latent} which monitors the ratio of weights flipped at each step. It is defined as:

\begin{equation} \label{eq:Pi_t}
    \pi_t = \log{\left( \frac{\text{\# flipped weights at time } t}{\text{Total number of weights}} + e^{-9} \right)}
\end{equation}

We used this evaluation parameter for analyzing the effects of each of the hyperparameters $\gamma, \sigma$ and $\tau$.

  \subsection{Second Order Binary Optimizer (Bop2ndOrder)} In this section, we present the complete ablation studies, comparisons, and tests on CIFAR10 and ImageNET of our optimizer.   
  
    \subsubsection{Biased or Unbiased. Batch or Layer Normalization}
\begin{figure}
      \centering
      \includegraphics[width=0.85\linewidth]{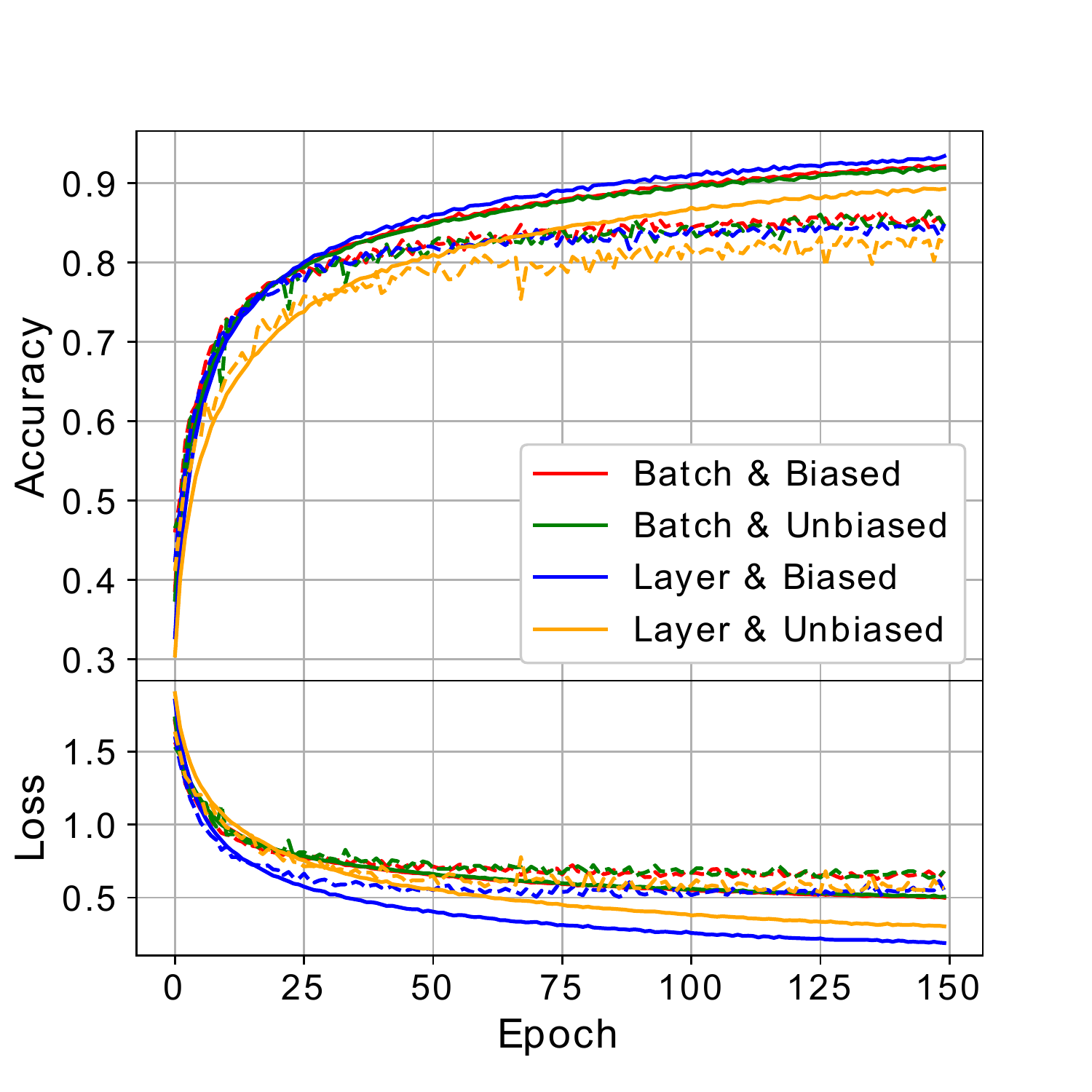}
      \caption{Bop2ndOrder tested with Batch Normalization and Layer Normalization combined with the Biased and Unbiased algorithms using the hyperparameters \\ $\gamma = 1e-7$, $\sigma = 1e-3$, $\tau = 1e-6$ and a batch size of 100. The line (-) refers to the training and the dashed line (- -) refers to the validation}
      \label{fig:BatchLayer_bop2ndorder}
    \end{figure}

    As previously stated, the Second Order Bop can be formulated in an unbiased or a biased case. Batch and Layer Normalization can also be used. In order to test this, in Figure \ref{fig:BatchLayer_bop2ndorder} we present a combination of the 4 possible cases. For choosing the correct hyperparameters, we did a search over all possible combinations of powers of 10 of each of the hyperparameters arriving to the combination $\gamma = 1e-7, \sigma = 1e-3, \tau = 1e-6$.
    
    In the original work \cite{Helwegen2019latent}, they used Batch Normalization (BN) in all of the tested architectures. Recently, Bethge et al \cite{Bethge2019simplicity} compared this layer against Layer Normalization (LN) in their Binary Dense Net architecture obtaining better results with LN. Also, Nayak et al \cite{Nayak2020} tested this idea using the Bop optimizer with the BinaryNet architecture in CIFAR10 obtaining similar accuracies but less cross-entropy loss with LN.
    
    As shown in Figure \ref{fig:BatchLayer_bop2ndorder}, the best results for the validation accuracy (top-1) are obtained when combining the BN with the unbiased version of the algorithm. As a matter of fact, the accuracy values for each experiment are not as far apart from each of the cases. The real issue is that using LN introduces lower losses without impacting the accuracy.

    \subsubsection{Hyperparameters exploration}
    To understand the effect of each of the three hyperparameters, we performed an ablation study based upon the optimal values that we previously obtained. Also, we implemented different schedulers for each of the hyperparameters.
    
      \textbf{Ablation Studies.} \\
      Taking as the starting point the ``optimal" values that we obtained for CIFAR-10 (100 epochs, BinaryNet, batch size of 100), we tested two magnitudes of 10 lower and two higher for each hyperparameter while letting the values of the other two remain constant.
      
\begin{figure*}
         \centering
         \begin{subfigure}[b]{0.33\linewidth}
             \centering
             \includegraphics[width=\linewidth]{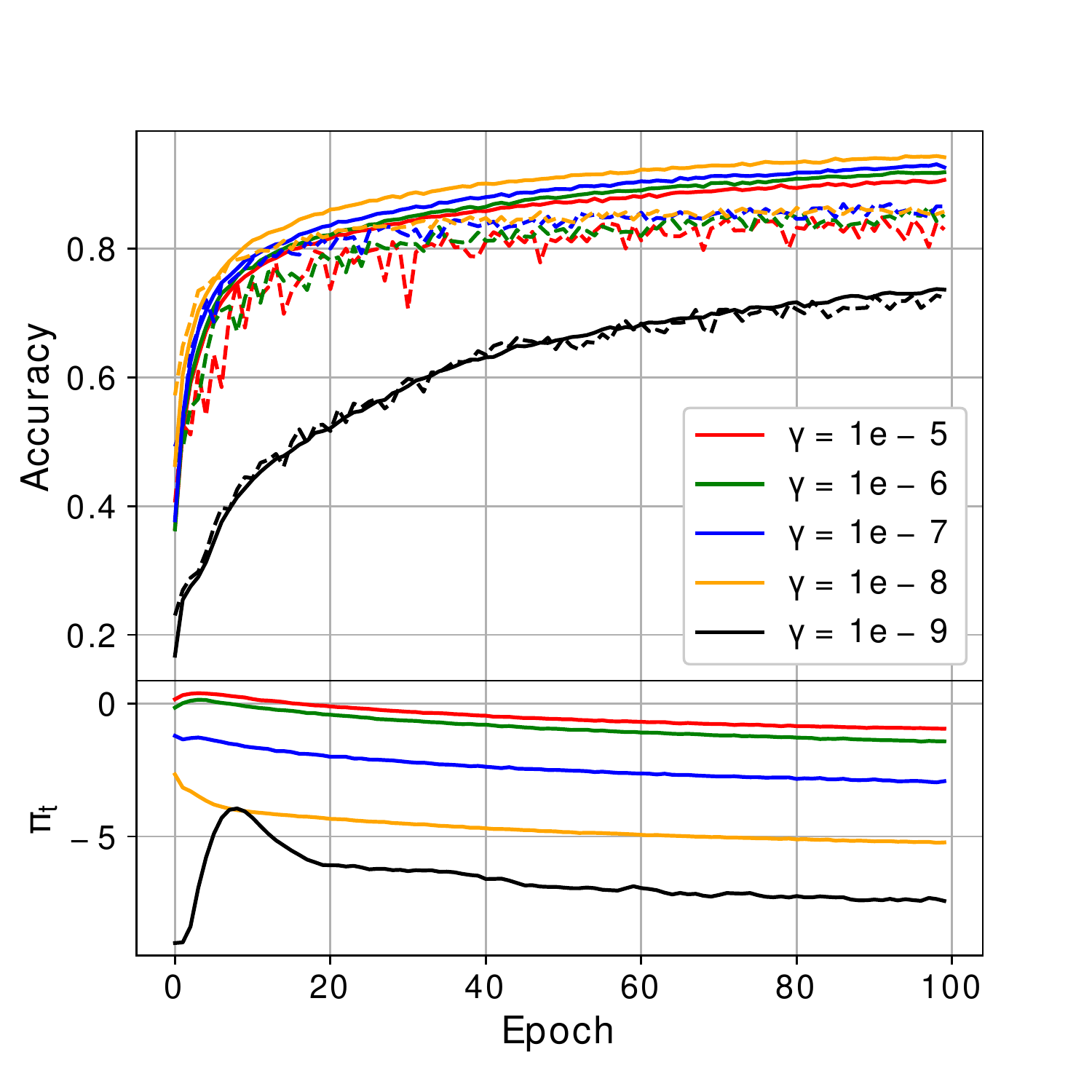}
             \caption{Testing different $\gamma$ values leaving \\ $\sigma = 1e-3, \tau = 1e-6$}
             \label{fig:gamma_tests}
         \end{subfigure}
         \hfill
         \begin{subfigure}[b]{0.33\linewidth}
             \centering
             \includegraphics[width=\linewidth]{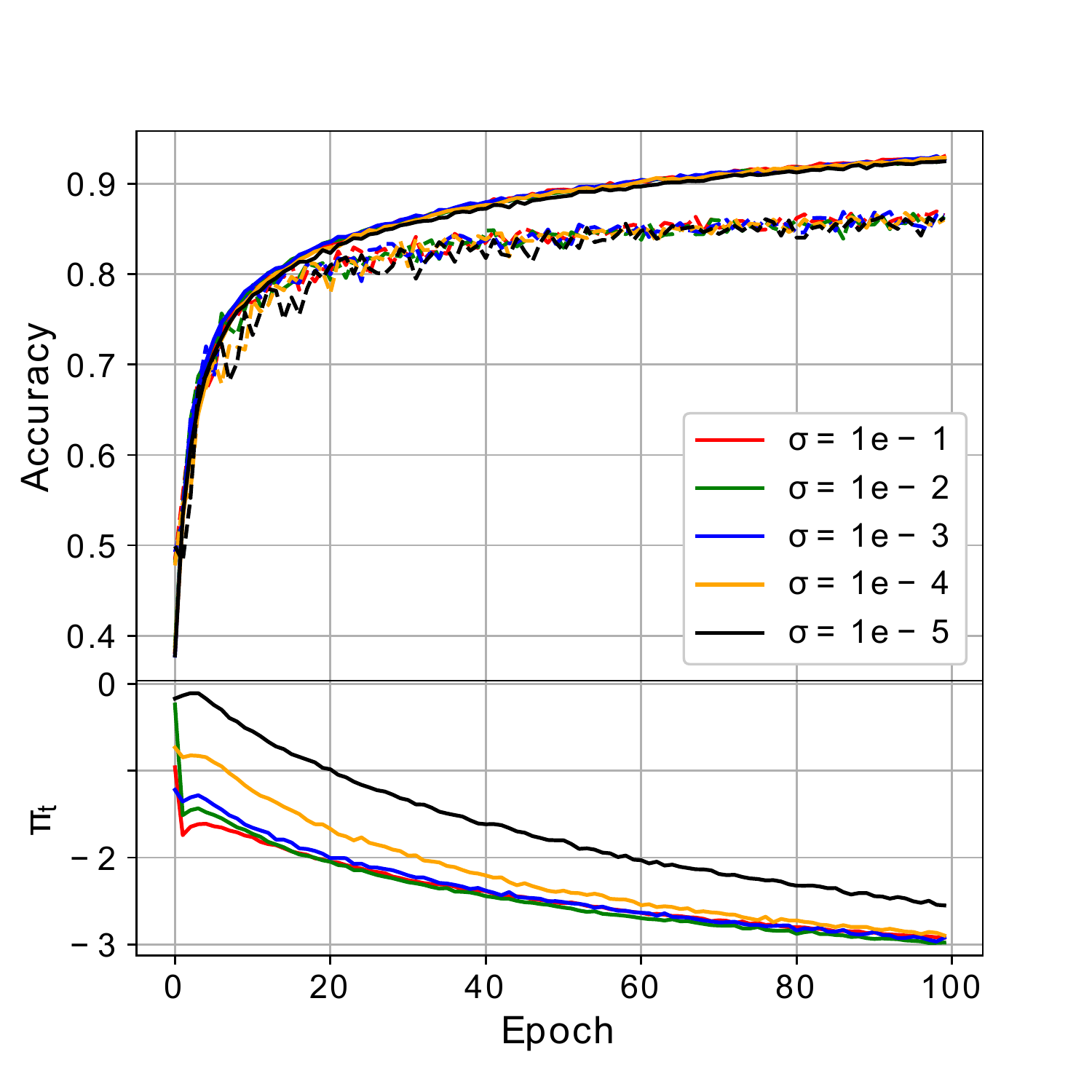}
             \caption{Testing different $\sigma$ values leaving \\ $\gamma = 1e-7, \tau = 1e-6$}
             \label{fig:sigma_tests}
         \end{subfigure}
         \hfill
         \begin{subfigure}[b]{0.33\linewidth}
             \centering
             \includegraphics[width=\linewidth]{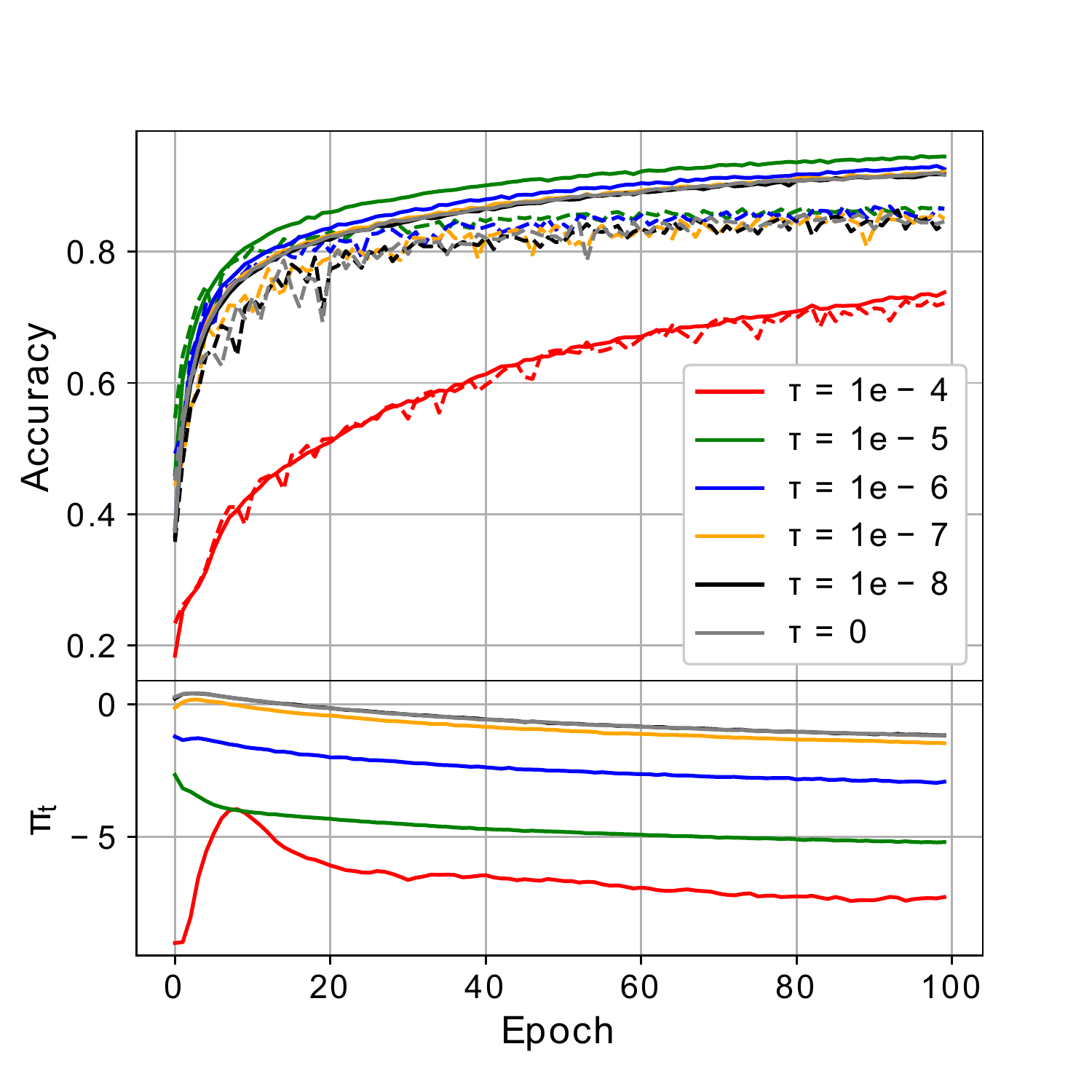}
             \caption{Testing different $\tau$ values leaving \\ $\gamma = 1e-7, \sigma = 1e-3$}
             \label{fig:tau_tests}
         \end{subfigure}
            \caption{Results of testing diverse hyperparameter values for Bop2ndOrder (unbiased) in BinaryNet for 100 epochs.}
            \label{fig:hyper_tests}
    \end{figure*}

      In figure \ref{fig:hyper_tests}, we show the results of this exploration. $\gamma$ and $\tau$ have similar effects to the learning rate (as expected from Bop \cite{Helwegen2019latent}). $\sigma$ works differently as its impact on accuracy is negligible. Its effect relies on increasing or decreasing the number of bit-flips. This could be used for stabilizing the network after some epochs by decreasing the value of this hyper-parameter.
      
\begin{figure*} [!h]
         \centering
         \begin{subfigure}[b]{0.4\linewidth}
             \centering
             \includegraphics[width=\linewidth]{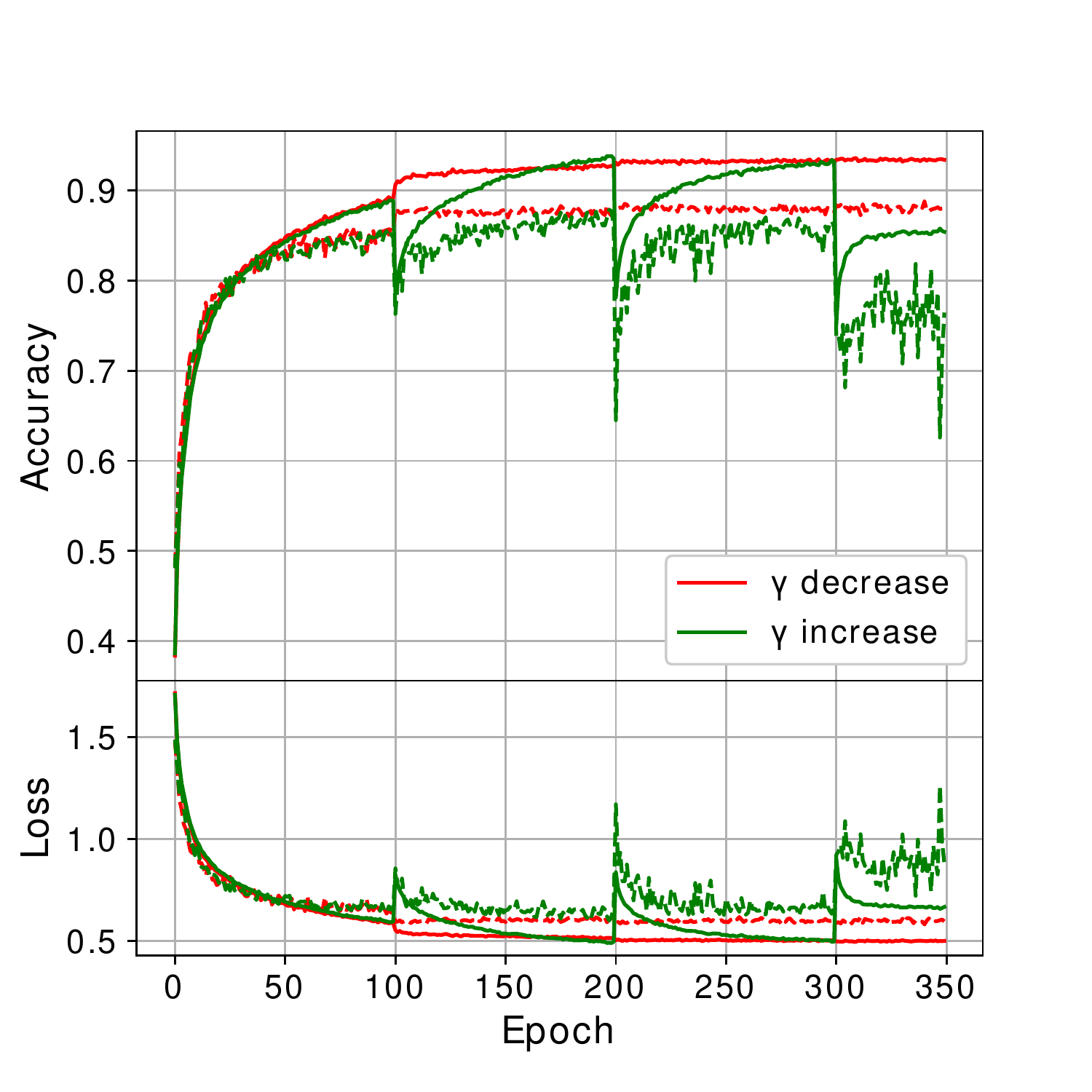}
             \caption{Exponential Scheduler (by powers of 10) applied to $\gamma$ in the unbiased version of the algorithm.}
             \label{fig:Unbiased_gamma_sched}
         \end{subfigure}
         \hfill
         \begin{subfigure}[b]{0.4\linewidth}
             \centering
             \includegraphics[width=\linewidth]{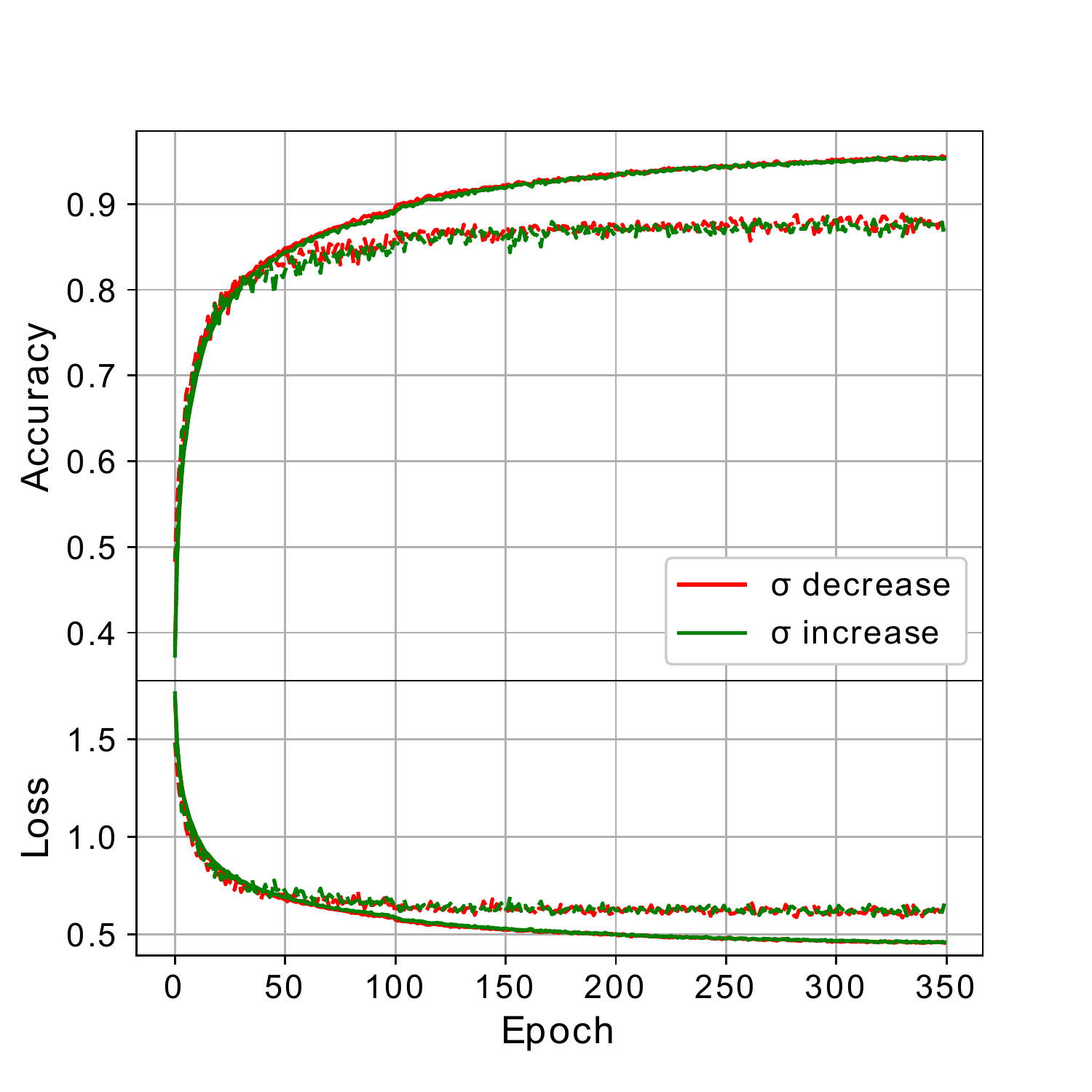}
             \caption{Exponential Scheduler (by powers of 10) applied to $\sigma$ in the unbiased version of the algorithm.}
             \label{fig:Unbiased_sigma_sched}
         \end{subfigure}
         \hfill
         \begin{subfigure}[b]{0.4\linewidth}
             \centering
             \includegraphics[width=\linewidth]{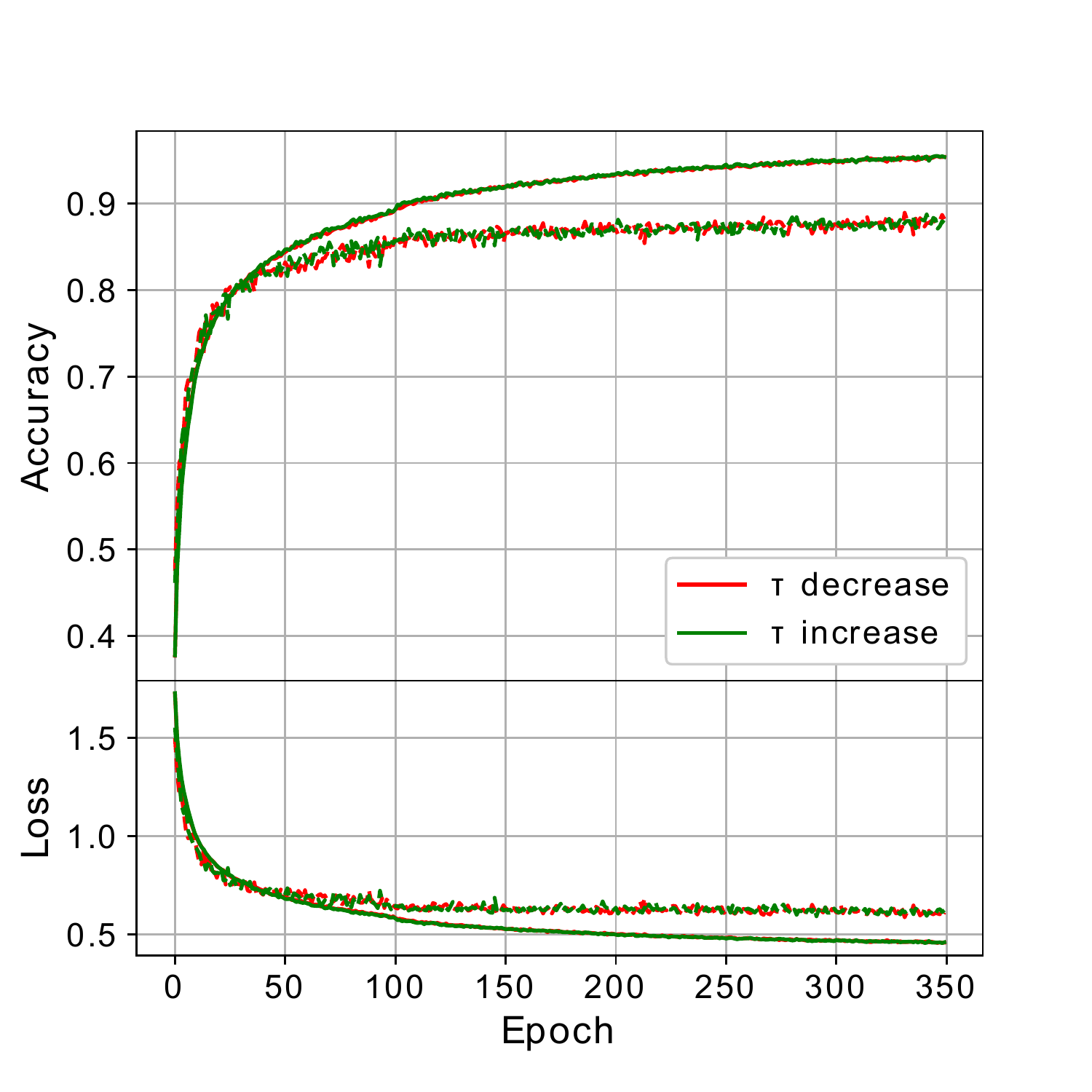}
             \caption{Exponential Scheduler (by powers of 10) applied to $\tau$ in the unbiased version of the algorithm. Unbiased algorithm. The scheduler does not change the behavior.}
             \label{fig:Unbiased_tau_sched}
         \end{subfigure}
         \hfill
         \begin{subfigure}[b]{0.4\linewidth}
             \centering
             \includegraphics[width=\linewidth]{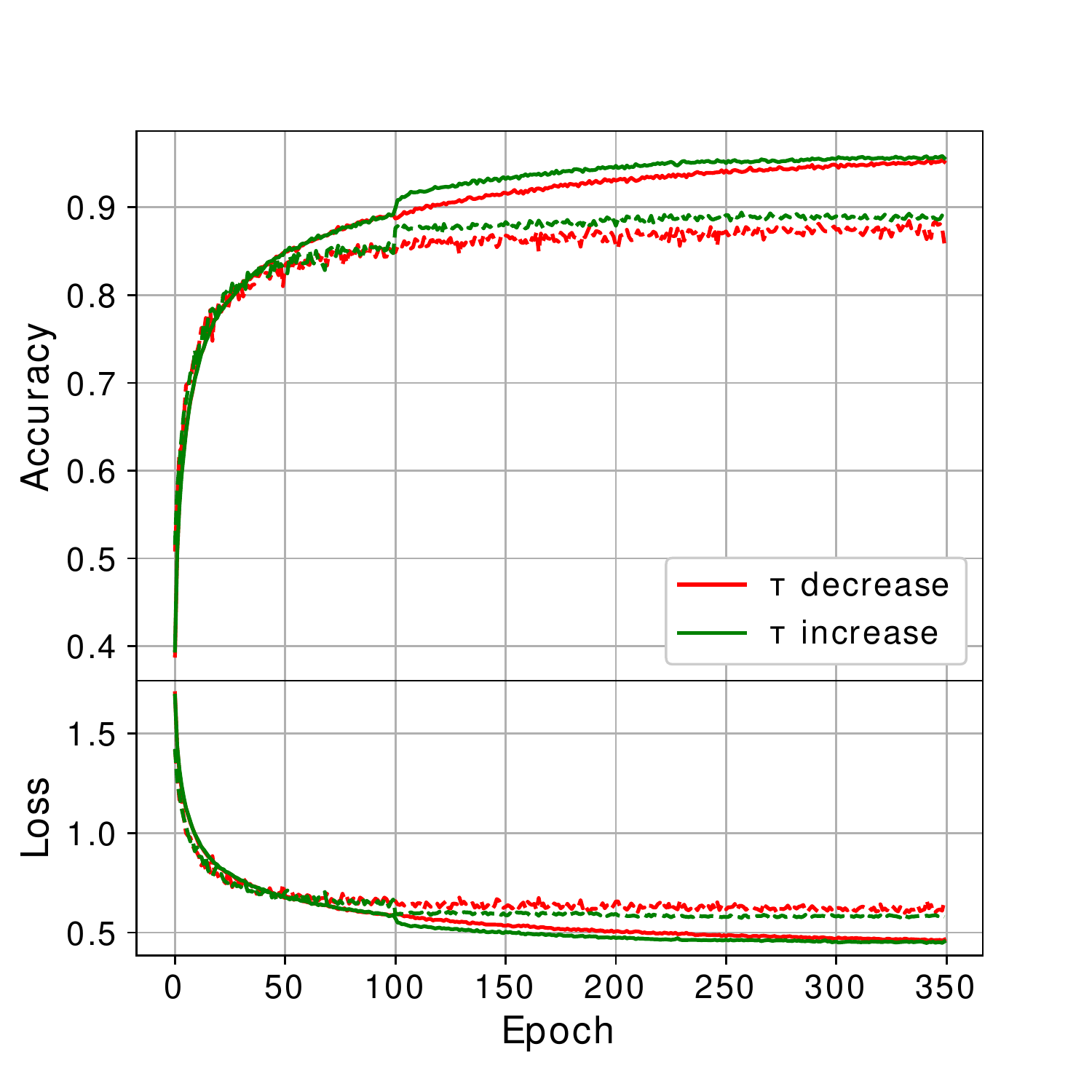}
             \caption{Exponential Scheduler (by powers of 10) applied to $\tau$ in the unbiased version of the algorithm. Biased algorithm. Here, the scheduler does change the behavior.}
             \label{fig:Biased_tau_sched}
         \end{subfigure}
            \caption{Exponential schedulers applied to the optimal hyperparameters of Bop2ndOrder unbiased and biased. For $\gamma$ and $\sigma$ the behavior is the same for both. For $\tau$, we have different effects on both algorithms.}
            \label{fig:sched_tests}
    \end{figure*}

      \textbf{The effect of schedulers.} \\
      In order to get a better grasp of the behaviour caused by these hyperparameters, we tested exponential schedulers (both increasing and decreasing by factors of 10 every 100 epochs show in figure \ref{fig:sched_tests}), for 350 epochs for the biased and the unbiased versions of the algorithm. For the base values, we chose the ones previously obtained $\gamma = 1e-7, \sigma = 1e-3, \tau = 1e-6$,
      
      In figure \ref{fig:sched_tests}, we show the results of applying these schedulers in the unbiased version of the algorithm. As expected, decreasing or increasing $\gamma$ directly affects how well the network learns. Decreasing it affects positively while increasing the value causes the algorithm to behave worse for a period of time before trying to set to the new value. For $\sigma$ both scheduling policies affect the hyperparameter equally; thus, it seems to be only a normalizing parameter. Lastly, $\tau$ seems to not affect the accuracy in a major way; this is counter-intuitive as having a higher value should avoid more weights to flip values and vice versa. Thus, we decided to explore the same schedulers with the biased algorithm. The behavior remains largely the same (for $\gamma$ and $\sigma$), but $\tau$ does affect the behavior of the algorithm; increasing the value of the threshold positively affects the accuracy. This could be explained by thinking of $\tau$ as a regularizer value by restricting or enabling more weights to be flipped. Thus, we want the algorithm to be more stable towards the end by increasing the value of the threshold.

    With these results, we can hypothesize that the biased version is more akin to be tuned by the hyperparameters than the unbiased version. Thus, to fine-tune the network, the biased algorithm should work better.

    \subsubsection{Bop vs. Bop2ndOrder}
\begin{figure} [!h]
      \centering
      \includegraphics[width=0.85\linewidth]{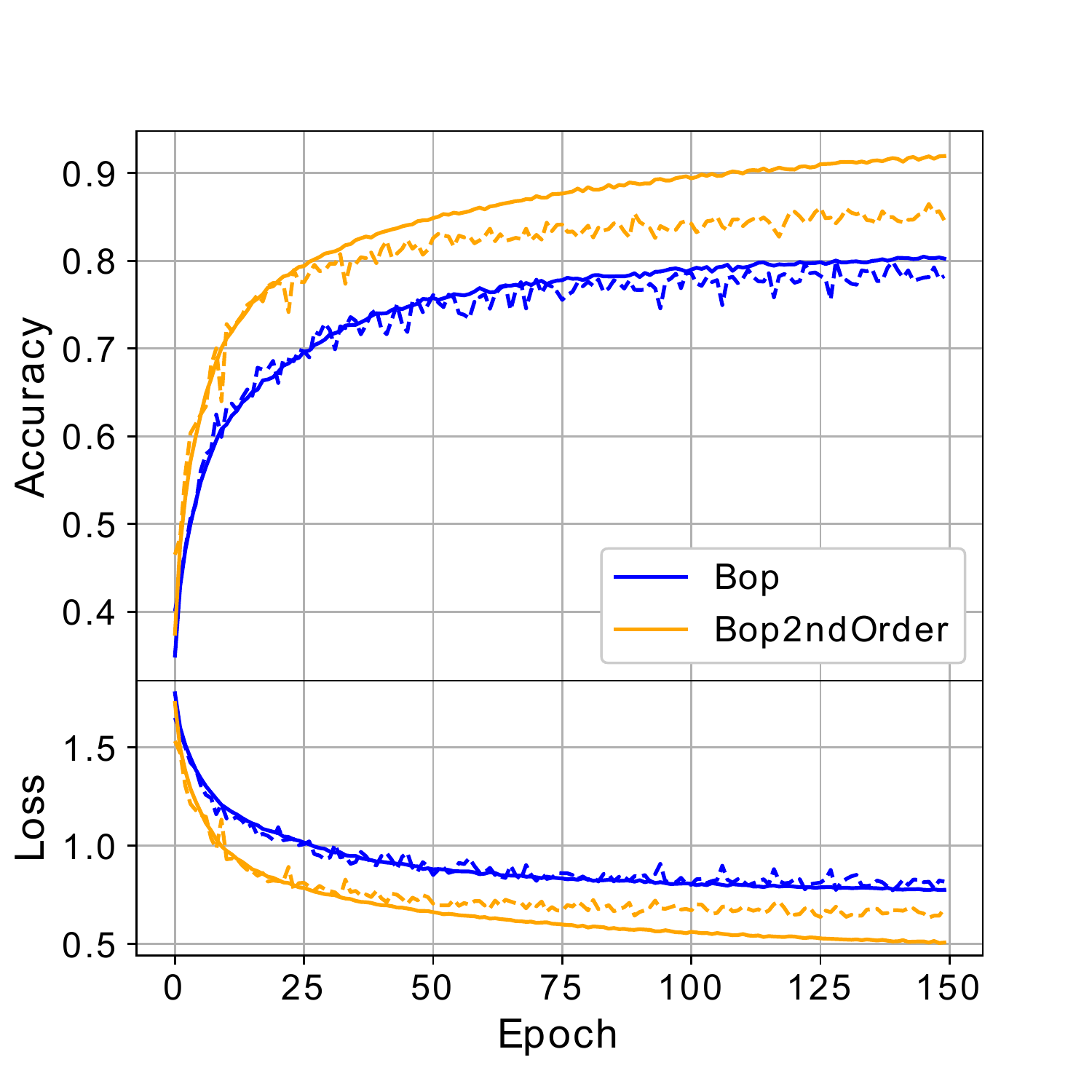}
      \caption{Bop and Bop2ndOrder compared in CIFAR10 using BinaryNet trained for 150 epochs. The line (-) refers to the training and the dashed line (- -) refers to the validation}
      \label{fig:comparison_vs}
    \end{figure}

    Bop2ndOrder has doubled the full-precision values as it is storing the second order momentum; however, the training times are increased by 15\%. 
    Considering this, we compared both optimizers (with their optimal hyperparameters). This is shown in Figure \ref{fig:comparison_vs}. As it can be seen, Bop2ndOrder is marginally superior with a validation accuracy (top-1) of \textbf{85.6\%} against the Bop accuracy of \textbf{79.6\%}. Also, it can be seen that Bop2ndOrder tends to overfit (Adam \cite{kingma2014adam} exhibits the same problem). Thus, a threshold increasing scheduler was used as a mean of coping with this issue.
    
    \begin{figure} [!h]
      \centering
      \includegraphics[width=0.85\linewidth]{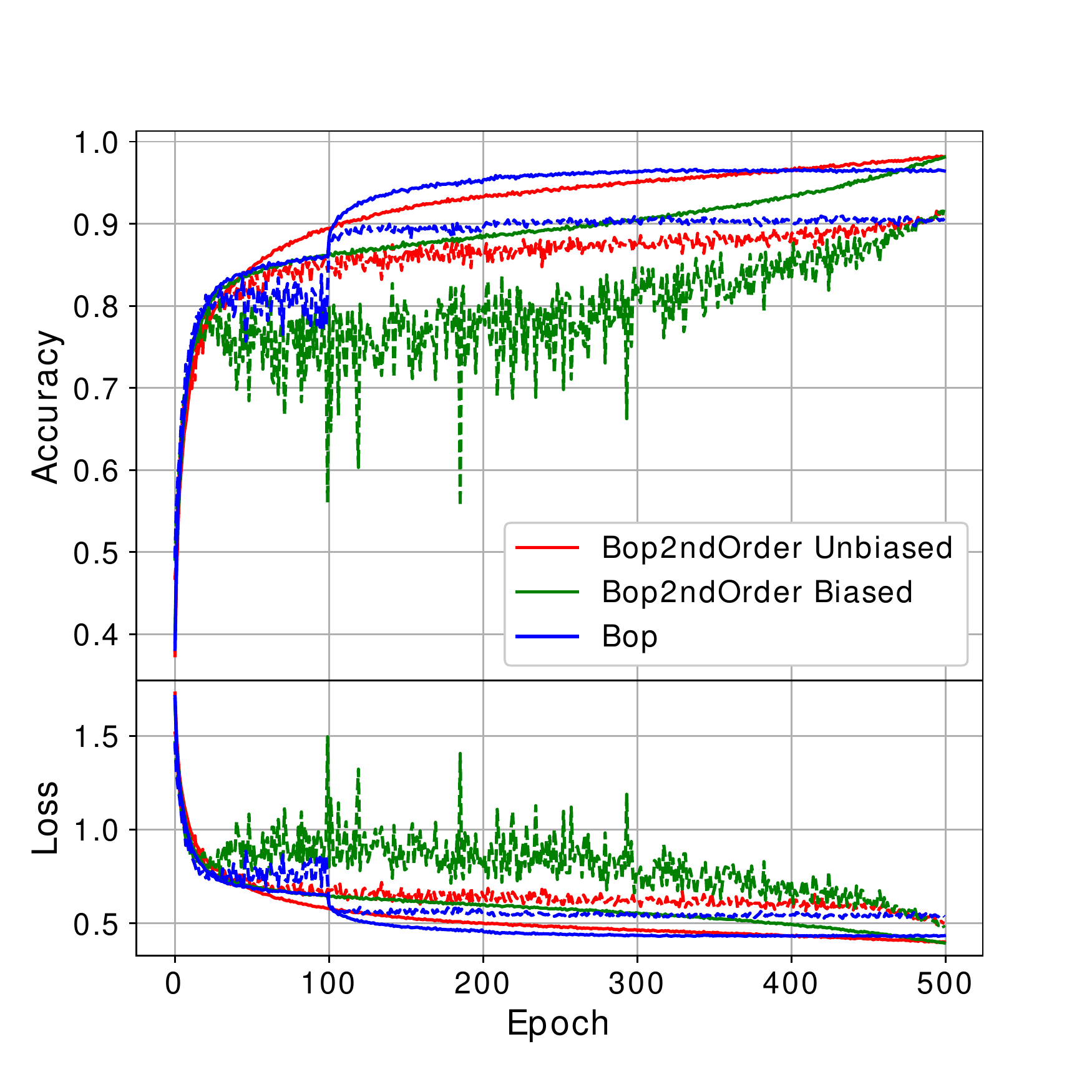}
      \caption{Optimum CIFAR10 run using Bop2ndOrder (biased and unbiased) compared to the Bop run. The line (-) refers to the training and the dashed line (- -) refers to the validation}
      \label{fig:cifarOpt}
    \end{figure}
    
     \begin{figure} [!h]
      \centering
      \includegraphics[width=0.85\linewidth]{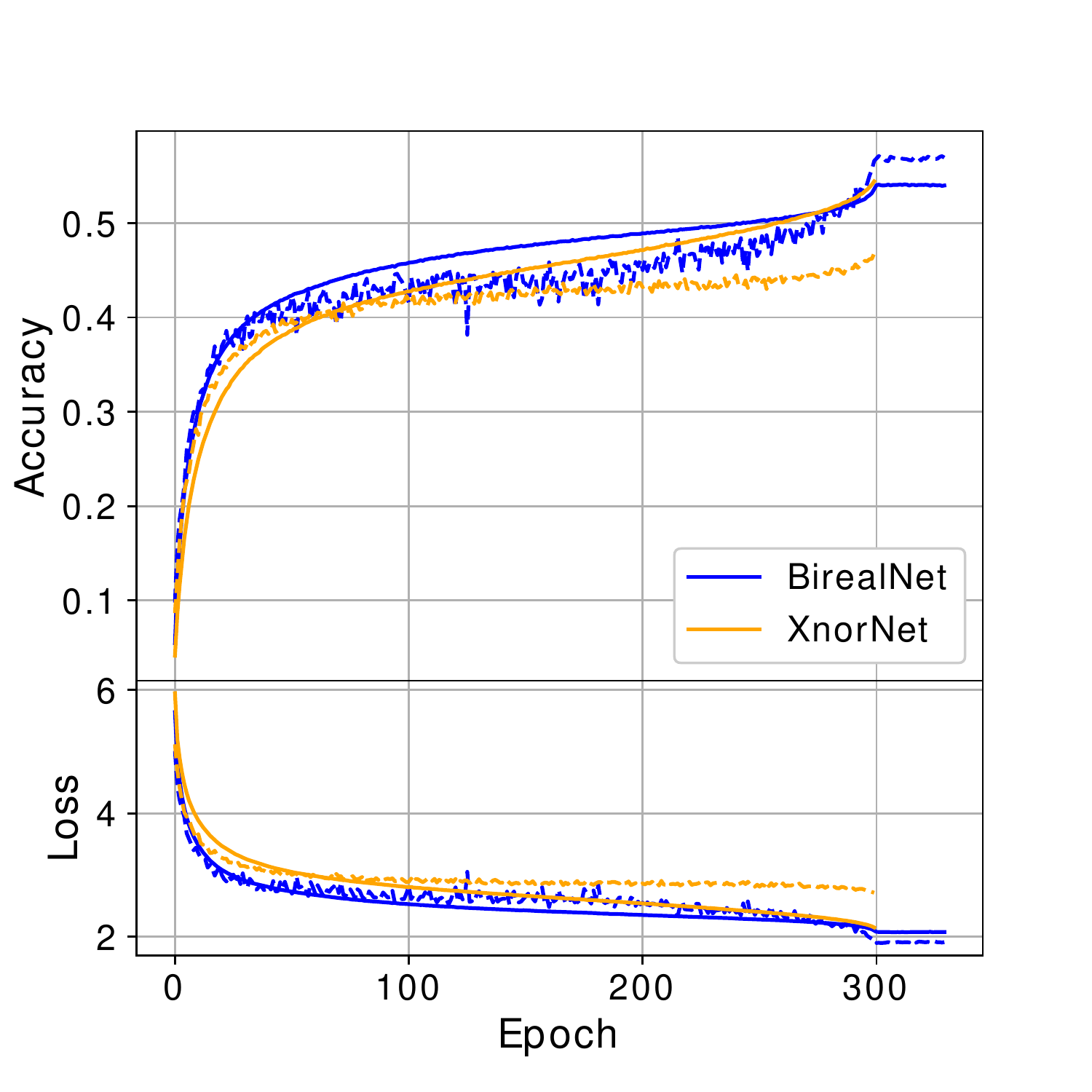}
      \caption{Optimum ImageNet using Bop2ndOrder and both the XnorNet and the BirealNet architectures. The line (-) refers to the train accuracy and the dashed line (- -) refers to the validation one}
      \label{fig:imagenet_XnorNet_birealnet}
    \end{figure}

  \subsection{CIFAR-10}
  In order to carry out a fair comparison with Bop \cite{Helwegen2019latent}, we used the BinaryNet architecture running for 500 epochs and using a batch size of 50. The difference here is the type of schedulers used. In our case, we used a polynomial scheduler with $\gamma = 1e-5 \rightarrow 1e-8, \sigma = 1e-2 \rightarrow 1e-5, \tau = 1e-7 \rightarrow 1e-2$ and a learning rate for the Adam optimizer of 0.01 that goes to 0.001 (polynomially) and otherwise default settings ($\beta_1 = 0.9, \beta_2 = 0.999$ and $\epsilon = 1e-7$). Also, we run the experiments with both the biased and the unbiased version of the algorithms.
  
  In figure \ref{fig:cifarOpt}, we show the results of this run. The validation accuracy (top-1) obtained for the biased version was of \textbf{91.9\%} and \textbf{91.4\%} for the unbiased one. Both results are higher than the Bop result of \textbf{91.3\%}. Running the given code for the Bop algorithm we obtained and accuracy of \textbf{91.0\%}.
  
  Both of our results only reach the best values at the last iterations - which is a behavior also presented with ImageNet -, while the Bop algorithm reaches an stable result 100-150 epochs before that. Also, the unbiased version is much more steady while the biased version is erratic. The results are summarized in Table \ref{table:CIFAR10_comp}.
  
\begin{table}[h]
\begin{center}
\begin{tabular}{|c|c|c|}
\hline
Optimizer                                                                              & Training Acc                     & Validation Acc          \\ \hline
Bop \cite{Helwegen2019latent}                                                                                    & 96.7\%                           & 91.0\%                  \\ \hline
\multirow{2}{*}{\begin{tabular}[c]{@{}c@{}}Bop2ndOrder\\ Unbiased (ours)\end{tabular}} & \multirow{2}{*}{\textbf{98.3\%}} & \multirow{2}{*}{91.5\%} \\
                                                                                       &                                  &                         \\ \hline
\begin{tabular}[c]{@{}c@{}}Bop2ndOrder\\ Biased(Ours)\end{tabular}                     & 98.1\%                           & \textbf{91.9\%}         \\ \hline
\end{tabular}
\end{center}
\caption{CIFAR-10 comparison between optimizers using Binary-Net \cite{Courbariaux2016binarized}}
\label{table:CIFAR10_comp}
\end{table}
  
  
  \subsection{ImageNet}
  We tested Bop2ndOrder on ImageNet using the binarized networks: XnorNet \cite{Rastegari2016xnornet} and BirealNet \cite{Liu2018bireal}. We also tried training with the BinaryNet architecture \cite{Courbariaux2016binarized}, but the increase in memory usage caused by storing the second order values proved to be high enough not to be feasible in a personal computer. The tests were done in a computer with 8 NVIDIA Tesla-P100 GPUs.
  
  We trained XnorNet for 100 epochs and BirealNet for 150 epochs with a batch size of 1024, and standard preprocessing with random flips and resize. This is the same as in Bop \cite{Helwegen2019latent} for comparison purposes. 
  
  For the hyperparameters, we used polynomial schedulers with $\gamma = 1e-4 \rightarrow 1e-9, \sigma = 1e-5 \rightarrow 1e-2, \tau = 1e-8 \rightarrow 1e-5$ and a learning rate for the Adam optimizer of $2.5e-3$ that goes to $5e-6$ (polynomially), and otherwise default settings ($\beta_1 = 0.9, \beta_2 = 0.999$ and $\epsilon = 1e-7$). In the case of XnorNet, we also used the l2-regularization of $5e-7$ that was used in Bop \cite{Helwegen2019latent}. The obtained results are summarized in Table \ref{table:Imagenet_comp}.
  
\begin{table}[h]
    \begin{center}
    \begin{tabular}{|c|c|c|c|}
    \hline
    Optimizer                                                                     & \multicolumn{1}{l|}{Acc} & \multicolumn{1}{l|}{XnorNet \cite{Rastegari2016xnornet}} & \multicolumn{1}{l|}{BirealNet \cite{Liu2018bireal}} \\ \hline
    \multirow{2}{*}{\begin{tabular}[c]{@{}c@{}}Bop2ndOrder\\ (ours)\end{tabular}} & Top-1                    & \textbf{46.9\%}              & \textbf{57.2\%}                \\ \cline{2-4} 
                                                                                  & Top-5                    & 70.9\%                       & 79.5\%                         \\ \hline
    \multirow{2}{*}{Bop \cite{Helwegen2019latent}}                                                          & Top-1                    & 45.9\%                       & 56.6\%                         \\ \cline{2-4} 
                                                                                  & Top-5                    & 70.0\%                       & 79.4\%                         \\ \hline
    \multirow{2}{*}{\begin{tabular}[c]{@{}c@{}}Latent\\ weights\end{tabular}}     & Top-1                    & 44.2\%                       & 56.4\%                         \\ \cline{2-4} 
                                                                                  & Top-5                    & 69.2\%                       & 79.5\%                         \\ \hline
    \end{tabular}
    \end{center}
    \caption{ImageNet comparison between optimizers using two common BNNs.}
    \label{table:Imagenet_comp}
\end{table}
  
  To further analyze the behavior of our optimizer, we also trained both architectures for 300 epochs with the unbiased version. In Figure \ref{fig:imagenet_XnorNet_birealnet} (orange  line) the results for the run using XnorNet are shown. The obtained validation accuracies are: \textbf{46.9\%} (top-1) and \textbf{71.3\%} (top-5). Still, we get better results than Bop (1\% in accuracy).
  
  In Figure \ref{fig:imagenet_XnorNet_birealnet} (blue line) the results for the run using BirealNet are shown. The obtained validation accuracies are: \textbf{56.7\%} (top-1) and \textbf{79.4\%} (top-5). We have a 0.1\% advantage over Bop, almost anecdotical. 
  
  There is a interesting fact about the graph of BirealNet, at 300 epochs the trend is an ``exponential" growth; thus, we decided to freeze the hyperparameters and train for 30 epochs more. With these new epochs, the results are: \textbf{57.1\%} (top-1) and \textbf{79.6\%} (top-5). We have a 0.5\% advantage over Bop.
  
\section{Discussion}
  
Referring to the results with CIFAR10 (Table \ref{table:CIFAR10_comp}) both of our algorithms present better results. Also, the biased version has higher validation accuracy even though its behavior is more chaotic. This could be due to its more tunable capacity as it is more affected by the modification of the hyperparameters (as previously discussed).

With respect to the ImageNet results (Table \ref{table:Imagenet_comp}) again our algorithm present better results than both Bop (by \textbf{0.6\%} with BirealNet and \textbf{1.0\%} with XnorNet) and the latent weights algorithms.

Our algorithm presents a peculiar behavior in both datasets (CIFAR-10 and ImageNet 2012) where the occurs a sudden increase in accuracy during the last epochs of the training. This could be due to the optimizer suddenly escaping from a local minimum, but there is no certainty of the reason of this behavior since this trend did not continue after increasing the number of iterations.

\subsection{Robustness}
    While testing different configurations of the hyperparameters on BinaryNet trained for 100 epochs with CIFAR-10, we noticed that the hyperparameters can be changed in power of 10 or less without having great impact in the validation accuracy, at least for the unbiased algorithm. The only thing to take into account is that the proportion between hyperparameters must be maintained. This indicates that Bop2ndOrder is robust enough to perform almost optimally when the exact hyperparameters are now known (at least for these experiments).
    
\section{Conclusions and Future Work}
In this paper we presented a new algorithm based on the interpretation of the BNN training methods as latent-weights encoding inertia \cite{Helwegen2019latent}. Thus, we decided to take this approach (parallel to that of momentum) and offer a second order approach which also uses the second raw moment estimate akin to Adam \cite{kingma2014adam}. With this new optimizer, we have surpassed the state-of-the-art results of BinaryNet on CIFAR-10, and the same for XnorNet and BirealNet on ImageNet 2012.

In general, the results surpass those presented in previous methods for optimizing BNNs. The caveat is that Bop2ndOrder uses more memory and training time (between 15\% - 25\% for the biased version and between 20 - 32\% for the unbiased one) than the other binary optimizer, Bop. Instead of choosing one over the other, there could be the case that both are used similar to how for training full-precision CNNs one optimizer (normally Adam) is used before another with less overfitting (such as RMSprop). Here, as Bop2ndOrder achieves faster a higher accuracy, could be used for the first epochs, and then change to Bop in order to stabilize the network and get less overfitting.

One exciting development that we foresee in the near future for these type of optimizers, is the introduction of specialized regularizers for BNNs (acting on the binary weights). Thus, Bop2ndOrder would be highly improved by using these methods as the training accuracies are way beyond those obtained using previous attempts.

{\small
\bibliographystyle{ieee_fullname}
\bibliography{egbib}
}

\end{document}